\documentclass{article}
\usepackage{spconf,amsmath,graphicx}
\usepackage{xcolor}
\usepackage{multirow}
\usepackage[hyphens]{url}
\usepackage{comment}
\usepackage[normalem]{ulem}

\title{Joint Learning On The Hierarchy Representation for Fine-Grained Human Action Recognition}
\name{{\parbox{\linewidth}{\centering
Mei Chee Leong\textsuperscript{1}, Hui Li Tan\textsuperscript{1}, Haosong Zhang\textsuperscript{1,2}, Liyuan Li\textsuperscript{1}, Feng Lin\textsuperscript{2}, Joo Hwee Lim\textsuperscript{1,2}\\
\{leong\_mei\_chee, hltan\}@i2r.a-star.edu.sg, haosong001@e.ntu.edu.sg, \\
asflin@ntu.edu.sg, \{lyli, joohwee\}@i2r.a-star.edu.sg}}
\thanks{This research is supported by the Agency for Science, Technology and Research (A*STAR) under its AME Programmatic Funding Scheme (Project A18A2b0046).}}

\address{Institute for Infocomm Research (I\textsuperscript{2}R), A*STAR\textsuperscript{1}\\
School of Computer Science and Engineering, Nanyang Technological University, Singapore\textsuperscript{2}
}

\begin{document}
%
\maketitle

\begin{abstract}
Fine-grained human action recognition is a core research topic in computer vision. Inspired by the recently proposed hierarchy representation of fine-grained actions in FineGym and SlowFast network for action recognition, we propose a novel multi-task network which exploits the FineGym hierarchy representation to 
achieve effective joint learning and prediction for fine-grained human action recognition. The multi-task network consists of three pathways of SlowOnly networks with gradually increased frame rates for events, sets and elements of fine-grained actions, followed by  
our proposed integration layers for joint learning and prediction. It is a two-stage approach, where it first learns deep feature representation at each hierarchical level, and is followed by feature encoding and fusion for multi-task learning. Our empirical results on the FineGym dataset achieve a new state-of-the-art performance, with 91.80$\%$ Top-1 accuracy and 88.46$\%$ mean accuracy for element actions, which are 3.40$\%$ and 7.26$\%$ higher than the previous best results. 
\end{abstract}

\section{Introduction}
\label{sec:intro}

As a fundamental research topic in computer vision, action recognition has always been an important research problem. Driven by advancements in deep learning models, such as 2D and 3D convolutional neural network (CNN) models, record breaking accuracies have been achieved on state-of-the-art action recognition datasets, such as ActivityNet \cite{caba2015activitynet}, Kinectics \cite{kay2017kinetics} and Sports1M \cite{akarpathy2014sports1m}.
Nonetheless, fine-grained action recognition, where the appearances of action and background are very similar and the recognition relies more on the modeling of temporal dynamics, remains a challenging research problem. To address fine-grained action recognition, various fine-grained datasets \cite{goyal2017something, chung2020haa500, verma2020yoga, shao2020finegym} have been recently proposed. Particularly, FineGym \cite{shao2020finegym} is a gymnastic video dataset with structured temporal and semantics hierarchy, decomposing an action into coarse-to-fine event-level, set-level, and element-level annotations. 
The scale and diversity of FineGym poses significant challenges for fine-grained action recognition, involving not only recognition of complex dynamics of intense motions, but also differentiation in subtle spatial semantics. Hierarchy representation of fine-grained actions  provides discriminating and complementary semantic, contextual and motion information at different granularity levels. However, this information is not fully utilized for joint representation and recognition, as most of the existing methods still focus on learning and prediction of single level action class. 

In this paper, inspired by the FineGym hierarchy representation, we propose a multi-task network for effective joint representation and learning for fine-grained action recognition. First, we introduce a new deep architecture according to the three-level hierarchy for effective learning of multi-level representation of fine-grained actions. The structure of the new architecture is designed to effectively learn and represent the associations and constraints across different levels of fine-grained action representation. Second, we propose a compact network of joint learning layers with multi-task learning to achieve joint learning and prediction of the three-level action class of the fine-grained action. Our experimental results on the FineGym benchmark dataset have shown a clear improvement on the state-of-the-art performance.

\section{Related Works}
\label{sec:related_works}

There are two main frameworks for fine-grained action recognition. The first type is the 2D+1D framework \cite{wang2017temporal, zhou2018temporal, lin2019tsm, girdhar2017actionvlad} where a 2D CNN model is used for extracting spatial features, and is followed by a 1D model for temporal aggregation. TSN \cite{wang2017temporal} samples frames from segmented video, and then aggregates the scores of each segment for final prediction.
TRN \cite{zhou2018temporal} introduces a temporal relational reasoning module that allows aggregation of multi-scale temporal relations between frames. TSM \cite{lin2019tsm} proposes a temporal shift module that achieves 3D performance with 2D complexity by shifting some channels along the temporal dimension. ActionVLAD \cite{girdhar2017actionvlad} aggregates spatio-temporal features from two-stream networks by pooling across space and time streams. 
 
The second type of framework is based on the 3D CNN model \cite{tran2015learning, carreira2017quo, wang2018nonlocal, feichtenhofer2019slowfast}. 
I3D \cite{carreira2017quo} expands 2D CNN into 3D CNN by adding a temporal dimension on filters and pooling kernels to capture the spatio-temporal features. 
Non-local neural network \cite{wang2018nonlocal} computes weighted sum of features in space-time to capture long-range dependencies. SlowFast \cite{feichtenhofer2019slowfast} comprises a slow and fast pathway to capture spatial semantic at low frame rate and learn motion information at high frame rate. Lateral connections are used to aggregate the features from the two pathways. SlowOnly is a variant of SlowFast without the fast pathway. 
There are also hybrid framework that fuses 2D and 3D CNN in a network \cite{sun2015human, feichtenhofer2016convolutional, tran2018closer, leong2020semi, feichtenhofer2020x3d}.

In existing works, investigations mainly focus on 
recognizing each level of the fine-grained actions separately. Different from existing works, we investigate joint spatio-temporal representation, learning and prediction of the hierarchical action class in fine-grained action recognition.

\section{Proposed Method}
\label{sec:method}
To exploit the hierarchical action representation and make use of the complementary contexts and semantics across different levels, we propose a multi-task network that consists of three input pathways for each hierarchy level and a compact integration network to jointly learn and predict the coarse-to-fine action representations.

\begin{figure}[t]
\centering 
\includegraphics[width=0.95\columnwidth]{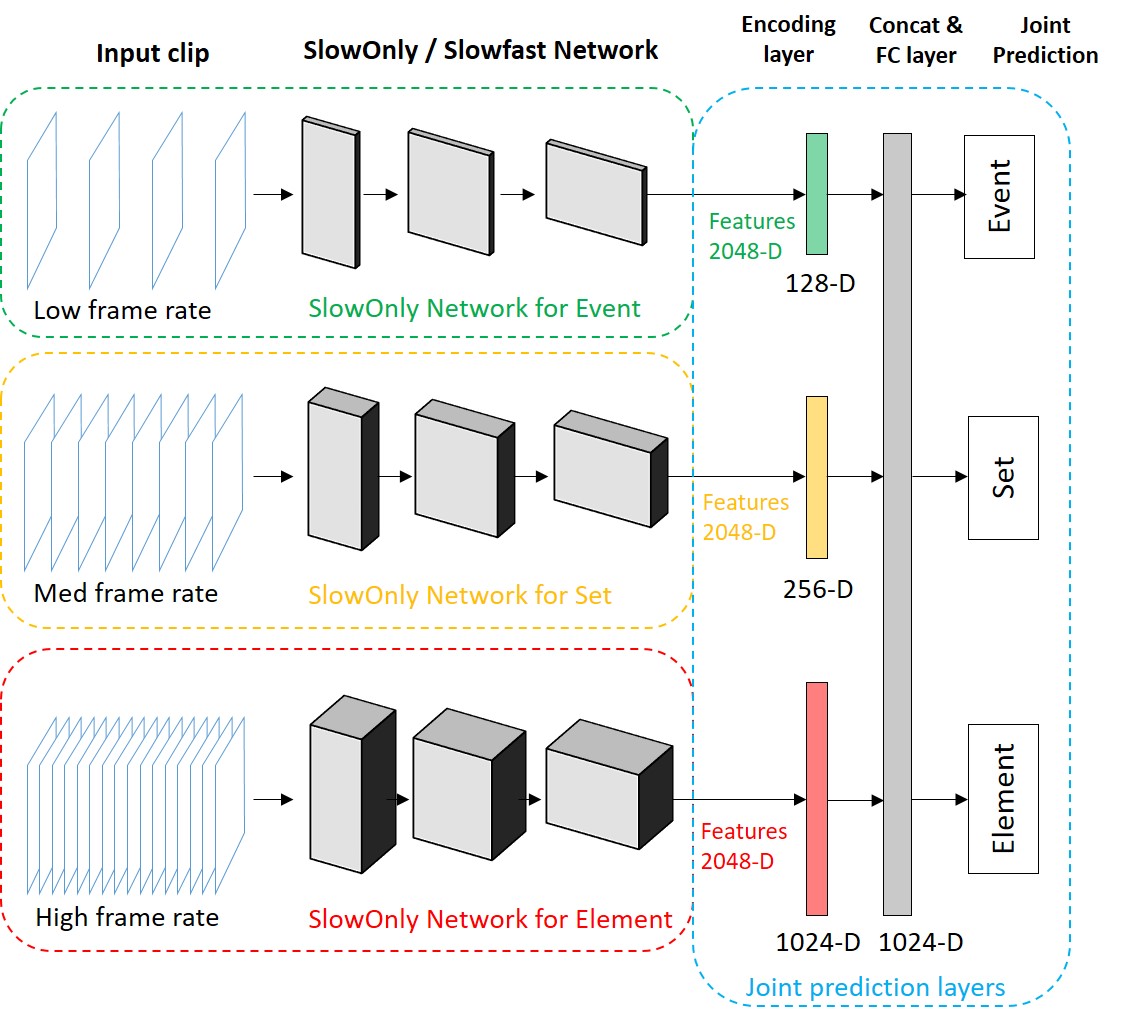} 
\caption{The architecture of our multi-task network for effective representation and joint learning of three-level hierarchy of fine-grained human actions.}
\label{fig:multi-task_features}
\end{figure}

\subsection{Hierarchy Representation of Fine-Grained Actions}
\label{ssec:representation}

In~\cite{shao2020finegym}, a hierarchy representation format is proposed to represent fine-grained human actions. The hierarchy tree is composed of three levels, {\em i.e.}, {\em events}, {\em sets}, and {\em elements}. The events represent the coarsest level of action categories. In FineGym, it represents the actions of different gymnastic routines. The sets represent middle level sub-action categories under the corresponding event node in the hierarchy tree. Each sub-action category represents a cluster of several technically and visually similar element actions. The elements represent the sub-action categories of the finest granularity under the corresponding set node in the hierarchy tree.

The hierarchy representation provides a multi-level coarse-to-fine representation of fine-grained human actions. There are several advantages of such representation for both learning and prediction. First, it represents a large set of fine-grained actions as a multi-level semantic hierarchy tree. The semantic hierarchy provides a domain constraint and solid foundation for joint learning and perception for coarse-level to fine-grained action understanding. 
Second, the hierarchy annotations are derived on a decision tree. Hence, the hierarchy representation encodes the distinctive features of semantic meaning, visual appearance and motion information on different levels. For example, some element actions can be distinguished from their background, objects and viewing angles, while some actions are mainly different in fine-grained motion features and duration. 
Training a single model on the element level actions may not be sufficient to learn the discriminative feature representation in a hierarchy structure. In addition, the hierarchy structure also encodes domain knowledge of action classes. Hence, exploiting multi-level action representations may help in better aggregation of spatial semantics and temporal dynamics, and learning of the domain class information implicitly. Such semantic information and domain knowledge might not be well exploited when training on a single level, especially the element-level for fine-grained action recognition.

\subsection{Network Architecture for Multi-task Learning}
\label{ssec:architecture}

Inspired by the SlowFast network that captures spatial and temporal semantics from its slow and fast pathways, we design a three-pathway multi-task network for joint learning of the hierarchical representation of fine-grained actions. Specifically, the network has Event pathway, Set pathway and Element pathway where each pathway samples the input clip at different frame rate and output features for our joint prediction layers. We first encode the features from each pathway to a lower dimensional vector, before fusing them for joint prediction. 
The encoded features from all pathways are concatenated for multi-task prediction on the hierarchical action representation. The architecture of the proposed network is illustrated in Fig.~\ref{fig:multi-task_features}, where the base model for each pathway is a SlowOnly network trained for single level prediction.

\subsubsection{Base model for individual pathways}
\label{ssec:base_model}
The observations in~\cite{feichtenhofer2019slowfast} suggest that, action representation such as the event class at the coarsest granularity is more related to visual appearance and background context, and it relies less on motion information. On the other hand, the element class at the finest granularity relies more on detailed motion information and has subtle differences in visual appearance~\cite{shao2020finegym}. According to the hierarchy representation, we propose to use a slow pathway of low frame rate for event class which captures semantic information from a few sparsely sampled frames, a medium pathway of moderate frame rate for set class which captures increasing motion information from increased sample frames, and a fast pathway of high frame rate for element class which captures the detailed motion information from densely sampled frames. Following the notation in~\cite{feichtenhofer2019slowfast}, we denote the feature shape of a pathway as $\{T, S^2, C\}$, where $T$ is the number of the frames sampled, $S$ represents the spatial size, and $C$ denotes the number of channels.

\subsubsection{Joint Prediction Layers}
\label{ssec:joint_pred}

After training the base models for each hierarchy level, we combine them to form the three pathways of our multi-task network. 
The pre-trained features for Event, Set and Element pathways are encoded from the output vector (2048-D) to a lower dimension vector of size 128, 256 and 1024 respectively. 
Element has the largest dimension of feature vector, and those of Set and Event are reduced gradually since more bits are required to encode the details of low-level sub-actions. 
This is to involve more temporal dynamic features extracted from high frame rate input, and sufficient spatial information from lower frame rate input. The features are then concatenated and full-connected to a linear layer of dimension 1024. Finally, the linear layer is connected to three classifiers for joint prediction of the event, set and element labels.
We adopt cross-entropy loss in (\ref{eq1}) for the three categories, where $c$ denotes category of event, set or element. $s_i^c$ is the CNN score of class $i$ of category $c$, $t_i$ is the ground-truth value and only one element $t_p$ is not zero for ground-truth positive class. $N^{c}$ denotes the number of classes in category $c$.
\begin{equation} \label{eq1}
L^{c} = -\sum_{i=1}^{N^c} t_i^c \log \left( \frac{e^{s_i^c}}{\sum_{j=1}^{N^c} e^{s_j^c}} \right) = - s_p^c + \log {\sum_{j=1}^{N^{c}}{e^{s_j^c}}}
\end{equation}
The total loss, $L^{total}$ in (\ref{eq2}), for training the multi-task network is a weighted sum of all losses, where the ratio for the weights ${\lambda_1}$, ${\lambda_2}$ and ${\lambda_3}$ are set as 1, 2 and 4 respectively.
\begin{equation}\label{eq2}
L^{total} = \lambda_1 L^{Event} + \lambda_2  L^{Set} + \lambda_3  L^{Element}
\end{equation}

\begin{table*}[!h]
\setlength{\tabcolsep}{5.5pt}
  \centering
  \small
  \caption{Individual model results for Event, Set and Element.}
    \begin{tabular}{|l|l|c|c|ccc|ccc|ccc|} \hline

    \multirow{2}{*}{Class} & \multirow{2}{*}{Network}  & \multirow{2}{*}{$T$} & \multirow{2}{*}{base $C$} & \multicolumn{3}{|c|}{Event} & \multicolumn{3}{|c|}{Set} & \multicolumn{3}{|c|}{Element}\\ \cline{5-13}

      &  & &  & Top-1 & Top-5 & Mean & Top-1 & Top-5 & Mean & Top-1 & Top-5 & Mean\\ \hline
    Event & SlowOnly & 4 & 64  & 99.28  & 100   & 99.22 &-&-&- &-&-&-   \\ \hline
    Set &  SlowOnly  & 8 & 64  &- &-&-& 95.49 & 99.95 & 95.43 &-&-&-   \\ \hline
    Set &  SlowOnly  & 16 & 64   &- &-&-& 97.70 & 99.83 & 97.68 &-&-&-   \\ \hline
    Element &  SlowOnly$^1$  & 16 & 64 &-&-&- &-&-&- & 79.30  &  - & 70.2  \\ \hline
    Element &  SlowOnly  & 32 & 64 &-&-&- &-&-&- & 91.05  &  97.82 & 88.40  \\ \hline
        Element & SlowFast  & 4 (slow) & 64 (slow) &-&-&- &-&-&- & 82.32 & 98.15 & 78.93 \\
     &   & 32 (fast) & 8 (fast) &&& &&&  & &  &    \\ \hline

    \end{tabular}%
  \label{tab:indv_results}%
\end{table*}%

\begin{table*}[htbp]
  \centering
  \small
  \setlength{\tabcolsep}{4.5pt}
  \caption{Multi-task network results for joint recognition.}
    \begin{tabular}{|l|l|l|ccc|ccc|ccc|} \hline
    \multicolumn{3}{|c|}{Combined pathways} &
 \multicolumn{3}{|c|}{Event} & \multicolumn{3}{|c|}{Set} & \multicolumn{3}{|c|}{Element}\\ \hline
        
     Event model & Set model & Element model & Top-1 & Top-5 & Mean & Top-1 & Top-5 & Mean & Top-1 & Top-5 & Mean\\ \hline
    SlowOnly & SlowOnly, $T=8$ & SlowOnly & 99.50 & 100 & 99.40 & 98.42 & 100 & 98.16 & 91.58 & 99.64 & 87.50   \\
    SlowOnly & SlowOnly, $T=16$ & SlowOnly & 99.54 & 100 & 99.37 & 98.94 & 100 & 98.87 & 91.80 & 99.69 & 88.46 \\
    SlowOnly & SlowOnly, $T=8$ & SlowFast & 99.81 & 100 & 99.63 & 98.18 & 100 & 97.75 & 82.68 & 98.51 & 77.62 \\ 
    SlowOnly & SlowOnly, $T=16$ & SlowFast & 99.81 & 100 & 99.49 & 98.78 & 99.98 & 98.47 & 82.95 & 98.73 & 77.76  \\ \hline

    \end{tabular}%
  \label{tab:combined_results_weighted_final}%
\end{table*}%

\begin{table}[!h]
  \centering
  \small
   \setlength{\tabcolsep}{1.5pt} 
  \caption{Comparison with State-of-the-Art performance on FineGym99 action recognition.}
    \begin{tabular}{|l|l|cc|cc|cc|} \hline
    \multirow{2}{*}{Model} & \multirow{2}{*}{Modality}  &  \multicolumn{2}{|c|}{Event} & \multicolumn{2}{|c|}{Set} & \multicolumn{2}{|c|}{Element}\\ \cline{3-8}
    
     &  & Top-1 & Mean & Top-1 & Mean & Top-1 & Mean\\ \hline
    TSN \cite{wang2017temporal} & 2Stream & 99.86 & 98.47 & 97.69 & 91.97 & 86.0 & 76.4     \\
    TRNms \cite{zhou2018temporal} & 2Stream & - & - & - & - & 87.8 & 80.2    \\
    TSM \cite{lin2019tsm} & 2Stream & - & - & - & - & 88.4 & 81.2    \\
    I3D \cite{carreira2017quo} & 2Stream & - & - & - & - & 75.6 & 64.4    \\
    NL I3D \cite{wang2018nonlocal} & 2Stream & - & - & - & - & 75.3 & 64.3    \\ \hline
    
    SlowFast & RGB & 99.81 & 99.63 & 97.70 & 97.40 & 79.16 & 74.93 \\
    
    Multi-task (Ours) & RGB & 99.54 & 99.37 & 98.94 & 98.87 & \textbf{91.80} & \textbf{88.46} \\ \hline
    \end{tabular}%
  \label{tab:sota}%
\end{table}%

\section{Experiments}
\label{sec:experiments}

\subsection{Implementation Details}
\label{ssec:implementation}

Our framework is implemented using MMAction2 ~\cite{2020mmaction2} utilizing four GPUs GeForce GTX2080Ti. We adopt ResNet-50 as the backbone for SlowOnly base models. The base models and multi-task network are trained with learning rate 0.01, momentum 0.9, weight decay $1e-4$ and gradient clip 40. Following the original configurations, SlowOnly is trained with learning rate decay at fixed step 90 and 110.

Our experiment is conducted on FineGym dataset \cite{shao2020finegym}, specifically Gym99 which contains annotation on 4 events, 15 sets, and 99 elements. At the point where we download the dataset, some of the videos are no longer available.

\subsection{Base Models}
\label{ssec:base}

Recently, there is a study\footnote{\url{https://github.com/open-mmlab/mmaction2/tree/master/configs/recognition/slowonly}} that implements SlowOnly network on FineGym99. With the parameters of $T=16$ and base $C=64$, 
the model produced 79.3\% Top-1 and 70.2\% mean accuracies, which are still lower than the state-of-the-art performance. We first implement and investigate training of SlowOnly networks individually for Event, Set and Element actions as our base models with different configurations. For Event prediction, we sample 4 frames with 16 frames interval as low frame rate input, i.e., $T=4$. 
For Set prediction, we experiment with two configurations, i.e., $T=8$ with interval 8, 
and $T=16$ with interval 4. 
For Element prediction, we sample 32 frames with interval 2, i.e., $T=32$. 
The size of the input image is $(224, 224)$. The first convolution kernel is set as $(1, 7 ,7)$ with base $C=64$.
Additionally, we also implement SlowFast network as a base model for Element prediction. The first convolution kernels are set as $(1, 7, 7)$ with $C=64$ and $(5, 7, 7)$ with $C=8$ for the slow and fast pathways respectively.
SlowFast follows the implementation of SlowOnly, but adopts Cosine Annealing learning rate schedule. Both SlowOnly and SlowFast models are trained for 120 epochs and tested on six clips with center crop. The results for the individual models, evaluated on Top-1, Top-5, and mean accuracies are shown in Table \ref{tab:indv_results}.  

As observed in Table \ref{tab:indv_results}, SlowOnly outperforms SlowFast network in Element prediction, with 8.7$\%$ gain in Top-1 accuracy. This is due to the reduced channel dimension in the fast pathway in SlowFast network which could not capture sufficient temporal context to distinguish the fine-grained motion. For Set and Element predictions, utilizing higher number of frame inputs leads to results improvement in Top-1 and mean accuracies. The results in Table \ref{tab:indv_results} are also served as baselines for comparison with our multi-task network.

\subsection{Multi-task Network}
\label{ssec:multi-task}

In this section, we take one base model from each hierarchy class in Table  \ref{tab:indv_results} to form the pathways of our multi-task network. During training, the weights of the base models are freezed, where we utilize the output features to train our joint prediction layers. The network is trained for 60 epochs and tested on a single clip with center crop. We experiment with four different configurations and obtain the results in Table \ref{tab:combined_results_weighted_final}.         

As observed in Table \ref{tab:combined_results_weighted_final}, multi-task networks with fused features perform better than their corresponding base model trained separately for each category 
in Table \ref{tab:indv_results}. This is due to the encoding and learning of complementary feature representation from multi-level spatio-temporal semantics. 
It is also apparent that multi-task networks with SlowOnly network performs better than SlowFast network in the Element prediction 
due to the difference in channels dimension.

\subsection{Comparison with State-of-the-Art}
\label{ssec:comparison}

We compare our multi-task learning results with state-of-the-art results of action recognition on FineGym99 presented in \cite{shao2020finegym}. Specifically, we compare with TSN \cite{wang2017temporal}, TRNms \cite{zhou2018temporal}, TSM \cite{lin2019tsm}, I3D \cite{carreira2017quo} and NL I3D \cite{wang2018nonlocal} with 2-stream modalities. The results are presented in Table \ref{tab:sota}. The Event, Set and Element results from TSN are trained separately, while our results are joint prediction from our multi-task network.  
Additionally, we implemented SlowFast with multi-task learning by modifying its last fully-connected layer to three classifiers and adopt the same weighted loss function for training.

Our network outperforms all the action models in Element prediction, achieving 91.80$\%$ in Top-1 accuracy and 88.46$\%$ in mean accuracy, with improvement of 3.40$\%$ and 7.26$\%$ respectively over state-of-the-art performance. For Set prediction, our joint learning results performs better than TSN, with 1.25$\%$ and 6.90$\%$ increment in Top-1 and mean accuracies respectively. 
As compared to the baseline SlowFast multi-task network, our network with encoding and fusion layers learns better joint representation that leads to improved performance in Element and Set predictions. For Event prediction, our performance is comparable to TSN and SlowFast results.

\section{Conclusion}
\label{sec:conclusion}
In this paper, we presented a multi-task network for effective representation and joint learning of fine-grained human actions on the three-level hierarchy proposed for FineGym. Our experiment results show the effectiveness of exploiting and leveraging on the semantic and temporal context of parallel pathways with varying input sampling. We added integration layers to allow joint encoding and learning of complementary spatio-temporal features of hierarchical action categories.  
Our multi-task network outperforms networks with single task prediction. For future work, we will look into networks with end-to-end training to jointly learn and refine the hierarchical action representations for multi-task prediction.

\bibliographystyle{IEEEbib}
\bibliography{refs}

\end{document}